\pdfoutput=1
\documentclass[11pt]{article}

\usepackage{acl}

\usepackage{times}
\usepackage{latexsym}

\usepackage[T1]{fontenc}
\usepackage[utf8]{inputenc}

\usepackage{microtype}

\usepackage[nolist]{acronym} % For acronyms
\usepackage{booktabs}
\usepackage{subcaption}
\usepackage{graphicx}
\usepackage{amsmath}
\usepackage{bbm}
\usepackage{multirow}
\usepackage{hyperref}
\usepackage{makecell}
\usepackage{romannum}
\usepackage[switch]{lineno}
\usepackage{soul,sidecap}

\title{Beyond Distributional Hypothesis: Let Language Models Learn\\ Meaning-Text Correspondence}

\author{
Myeongjun Jang$^1$~~~
Frank Mtumbuka$^1$~~~
Thomas Lukasiewicz$^{2,1}$~~~
\smallskip 
\\
$^1$\,Department of Computer Science, University of Oxford, UK \\
$^2$\,Institute of Logic and Computation, TU Wien, Austria \\
firstname.lastname@cs.ox.ac.uk \\
}

\begin{document}
\pagenumbering{arabic}

\maketitle

% \runningpagewiselinenumbers
% \linenumbers

\begin{abstract}
The \ac{LNP}, which implies generating different predictions for semantically opposite inputs ($p$ is true iff $\lnot p$ is false), is an important property that a trustworthy language model must satisfy. However, much recent evidence shows that large-size \acp{PLM} do not satisfy this property. In this paper, we perform experiments using probing tasks to assess \acp{PLM}' \ac{LNP} understanding. Unlike previous studies that only examined negation expressions, we expand the boundary of the investigation to lexical semantics. Through experiments, we observe that \acp{PLM} violate the \ac{LNP} frequently. To alleviate the issue, we propose a novel intermediate training task, named \textit{meaning-matching}, designed to directly learn a meaning-text correspondence, instead of relying on the distributional hypothesis. Through multiple experiments, we find that the task enables \acp{PLM} to learn lexical semantic information. Also, through fine-tuning experiments on 7 GLUE tasks, we confirm that it is a safe intermediate task that guarantees a similar or better performance of downstream tasks. Finally, we observe that our proposed approach\footnote{\href{https://github.com/MJ-Jang/beyond-distributional}{https://github.com/MJ-Jang/beyond-distributional}} outperforms our previous counterparts despite its time and resource efficiency. 
% Code and data are available
% here: \href{https://github.com/MJ-Jang/Londinium}{https://github.com/MJ-Jang/Londinium}.
\end{abstract}

\section{Introduction}
Contemporary large-size \acp{PLM}, such as BERT \cite{BERT}, ELECTRA \cite{electra}, and GPT-2 and -3 \cite{GPT2, GPT3}, have shown excellent results in many downstream tasks, even performing better than humans in the GLUE \cite{GLUE} and SuperGLUE \cite{SuperGLUE} benchmark datasets. 
%Of late, they have become the foundation of modern \ac{NLP} systems based on their outstanding performance.

However, their reliability is recently being challenged. Many studies have conducted various probing tasks and observed that \acp{PLM} exhibit faulty behaviours, such as insensitiveness to sentence ordering \cite{pham2020out, gupta2021bert, sinha2021unnatural}, incomprehension on number-related representations \cite{wallace2019nlp, lin2020birds, nogueira2021investigating}, and lack of semantic content understanding \cite{ravichander2020systematicity, elazar2021measuring}. These issues raise concerns about \acp{PLM}' stability and reliability, precluding them from applications in practice, especially in risk-sensitive areas.

Another critical problem of \acp{PLM} is their inaccurate behaviour on \textit{negation}, which is a principal property in many language understanding tasks. For tasks where the \ac{LNP} holds ($p$ is true iff  $\lnot p$ is false; see \citealt{aina2018distributional}), \acp{PLM} should make different answers for the original and negated inputs. However, several studies observed that \acp{PLM} violate this property. In masked knowledge retrieval tasks, \acp{PLM} frequently generate incorrect answers for  negated input queries \cite{ettinger2020bert, kassner2020negated}. In other studies, \acp{PLM} show a poor generalisation ability on negated \ac{NLI} datasets \cite{naik2018, hossain2020analysis}.

Although the aforementioned studies produced promising analysis results, they limited the scope of the \ac{LNP} only to adding negation expressions (e.g., ``no'' and ``not''). However, other perturbations that generate the opposite meaning also can be applied to the property. Therefore, a consideration of such perturbation methods is necessary to fully assess whether \acp{PLM} satisfy the \ac{LNP}.

Also, remedies to alleviate the problem have not been studied much yet. \citet{hosseini2021understanding} recently employed data augmentation and unlikelihood training \cite{welleck2019neural} to prevent models from generating unwanted words, given the augmented negated data during \ac{MLM}. However, this approach has several downsides. First, like previous works, \citet{hosseini2021understanding} only considered negation expressions. Second, the data augmentation method is contingent on many additional linguistic components, %such as Semgrex pattern \cite{Semgrex}, dependency parser and \ac{POS} tags,
which causes the dependency of a model's performance on certain modules and precludes applying the method to other languages where such resources are unavailable. Finally, the model should be pre-trained from scratch with the unlikelihood objective, which consumes considerable time and resources.

In this paper, we expand the boundary of the \ac{LNP} to lexical semantics, i.e., \textit{synonyms} and \textit{antonyms}, and ascertain that \acp{PLM} are prone to violate the \ac{LNP}. Next, we propose a remedy, called \textbf{i}ntermediate-training on \textbf{m}eaning-\textbf{m}atching ($\textrm{I}\textrm{M}^2$), which hardly employs additional linguistic components. We hypothesise that a leading cause lies in the \ac{MLM} training objective, which assumes the \textit{distributional hypothesis} for learning the meaning of the text \cite{sinha2021masked}. Instead, we design a model that directly learns the correspondence between  words and their semantic contents. Through experiments, we verify that our approach improves the model's comprehension of the \ac{LNP}, while showing a stable performance on multiple downstream tasks.

Our main contributions are as follows: 
(\romannum{1}) We extend the investigation of the \ac{LNP} from negation to lexical semantics (Section~\ref{sec:probing_tasks}), 
(\romannum{2}) we reveal that \acp{PLM} are prone to violate the \ac{LNP} (Section~\ref{section.probing_exp}),
(\romannum{3})~we propose a novel remedy, named $\textrm{I}\textrm{M}^2$, which is decoupled from the \textit{distributional hypothesis} but learns meaning-text correspondence instead (Section~\ref{sec:remedy_imm}), 
(\romannum{4})~through experiments, we ascertain that the proposed approach improves the understanding of negation and lexical semantic information (Sections~\ref{sec:mm_sar} and \ref{sec:mm_mkr_mwr}), and  
(\romannum{5}) we verify that meaning-matching is a stable and safe intermediate task that produces a similar or better performance in multiple downstream tasks (Sections~\ref{sec:glue_exp} and \ref{sec:negnli_exp}).

%%%%%%%%%%%%%%%%%%%%%%%%%%%%%%%%%%%%%%%%%%%%%%%%%%%%%%%%%%%%
\begin{figure*}[t]
	\centering
	\begin{subfigure}[b]{0.3\textwidth}
		\includegraphics[width=\linewidth]{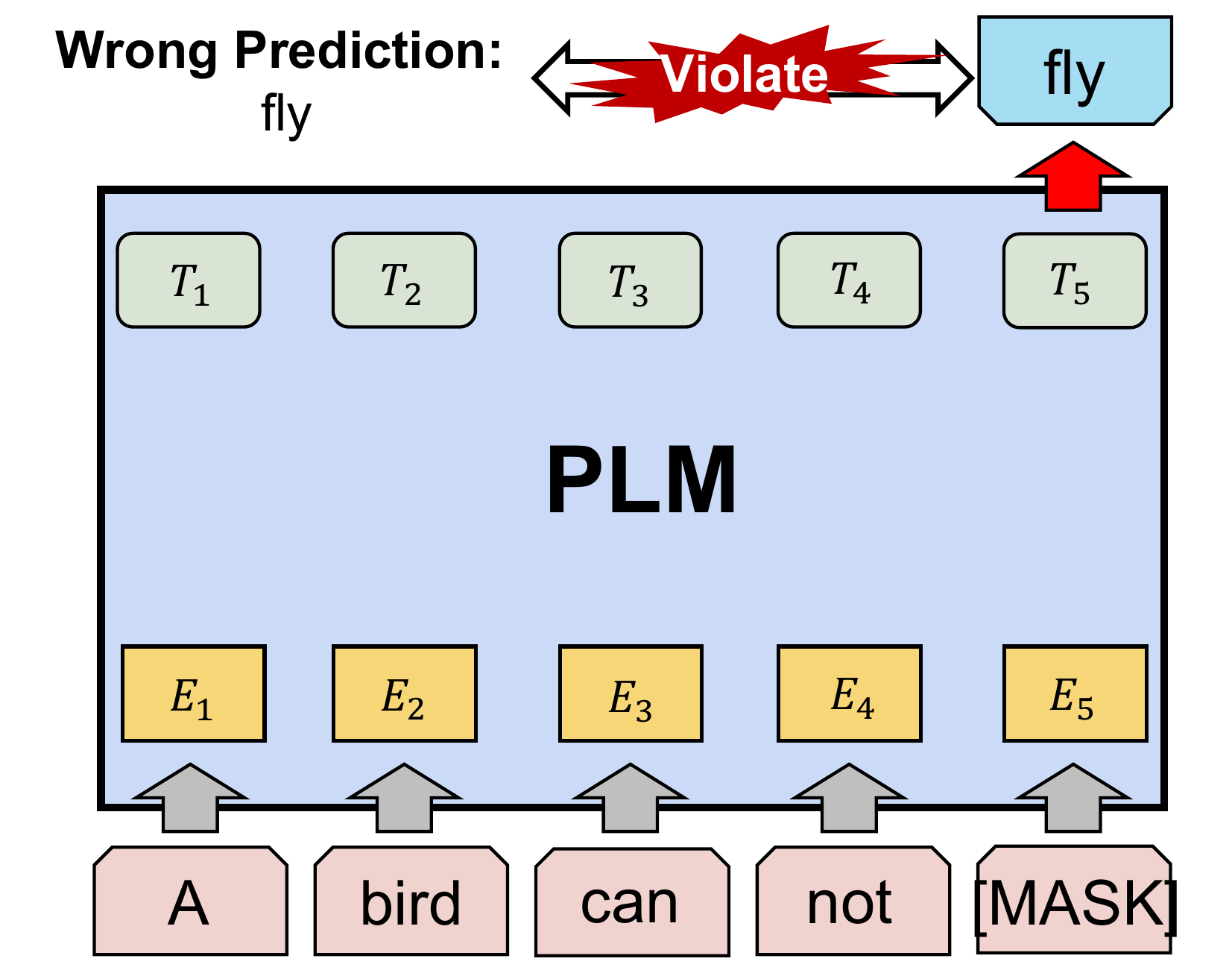}
		\vspace*{-2ex}
		\caption{MKR-NQ} \label{fig:mkr_nq}
	\end{subfigure} \
	\begin{subfigure}[b]{0.3\textwidth}
		\includegraphics[width=\linewidth]{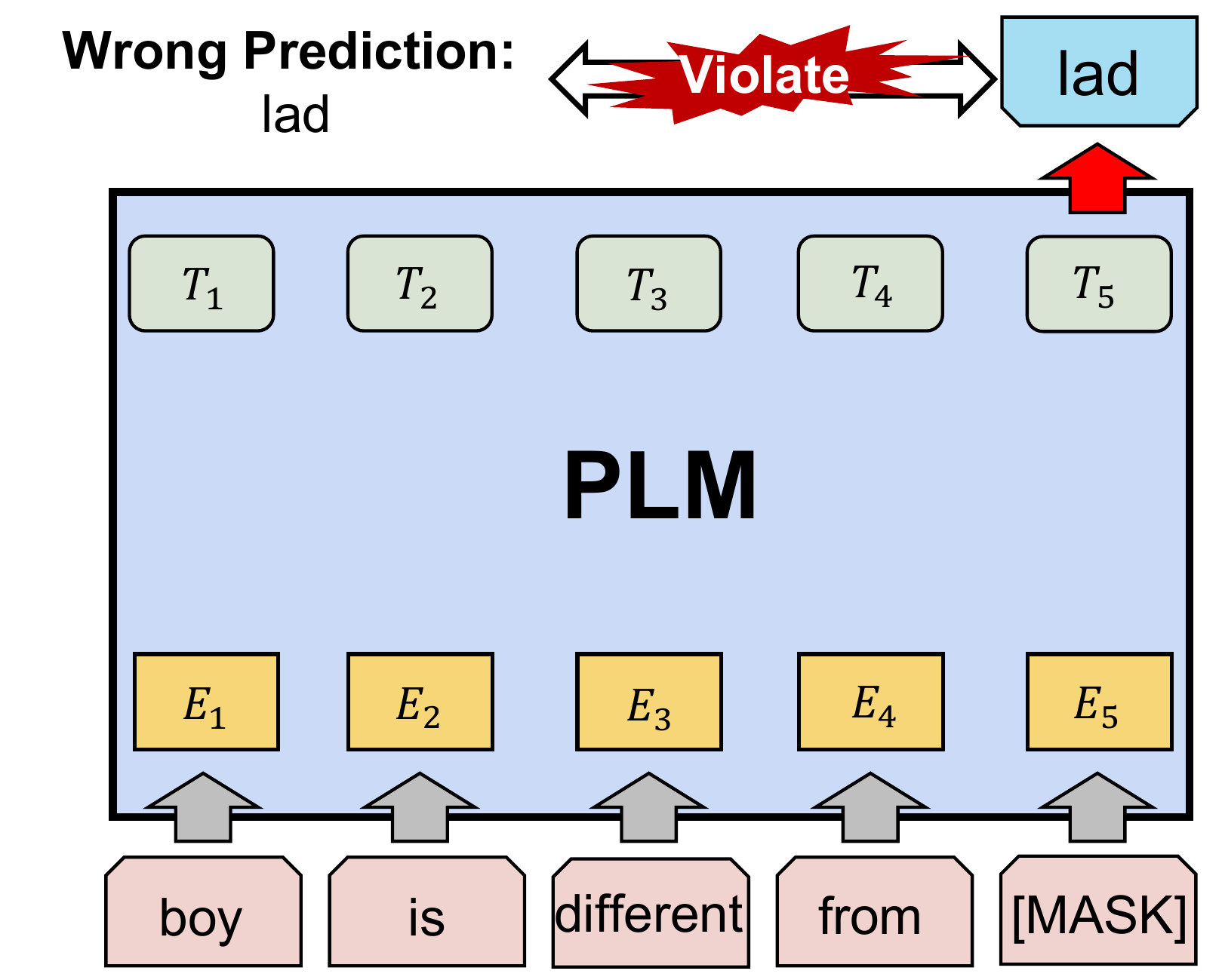}
		\vspace*{-2ex}
		\caption{MWR} \label{fig:mwr}
	\end{subfigure} \
		\begin{subfigure}[b]{0.3\textwidth}
		\includegraphics[width=\linewidth]{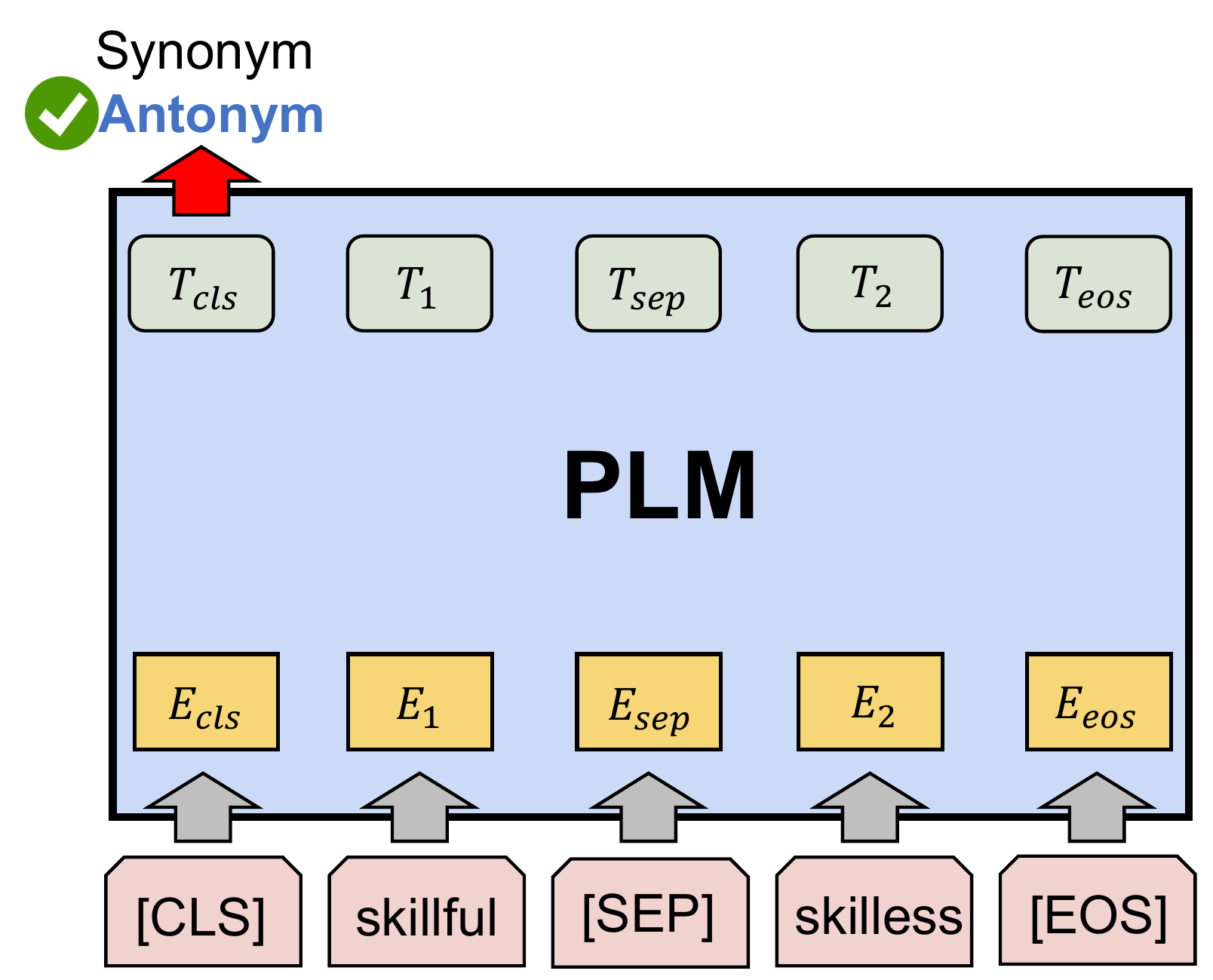}
		\vspace*{-2ex}
		\caption{SAR} \label{fig:sar}
	\end{subfigure} \
	\vspace*{-2ex}
	\caption{Illustration of the MKR-NQ, MWR, and SAR tasks.}
	\vspace*{-2ex}
	\label{fig:task_illustration}
\end{figure*}
%% -----------------------------------------------------
\vspace{-0.5ex}
\section{Probing Tasks for Investigating the Logical Negation Property}\label{sec:probing_tasks}
\vspace{-0.5ex}
We design three probing tasks to evaluate whether \acp{PLM} satisfy the \ac{LNP}: \ac{MKR-NQ}, \ac{MWR}, and \ac{SAR}. Brief illustrations of each task are  in Figure~\ref{fig:task_illustration}.

\subsection{Masked Knowledge Retrieval on Negated Queries}
The \ac{MKR-NQ} task examines whether \acp{PLM} generate incorrect answers for  negated queries. Following the work of \citet{kassner2020negated}, we constructed the evaluation dataset by negating the LAMA dataset \cite{petroni2019language}, which contains  masked free-text forms of ConceptNet \cite{conceptnet} triplets and their corresponding answers (e.g., (bird, CapableOf, fly) $\rightarrow$ (``A bird can [MASK]'', fly)). The task aims to generate a correct word through~\ac{MLM}.

According to the \ac{LNP}, a model must not generate the original answer if the query is negated. To measure how likely \acp{PLM} generate wrong predictions for negated queries, we collected pairs of (negated\_query, wrong\_predictions). We selected several relations in the LAMA dataset that ensure mutual exclusiveness between the original and negated queries.\footnote{For example, the \textit{HasProperty} relation is not suitable to use, because sentences like “Some adults are immature” and “Some adults are not immature” are not mutually exclusive.} For negating sentences, we selected LAMA data points that contain a single verb using the Spacy \ac{POS} tagger \cite{spacy}. Next, we added negation expressions, such as ``not'' and ``don't'', or removed such expressions if they existed. Finally, we collected the wrong predictions from ConceptNet by using the head entity and relation. As a result, we collected 3,360 data points for this task. The list of the relations that we used and examples of the data are  in Table~\ref{table.example_MKR_NQ} in Appendix~\ref{sec:example_appendix}. 

\subsection{Masked Word Retrieval}
To expand the boundary of the \ac{LNP} to lexical semantics, we design the \ac{MWR} task, which generates an answer of a masked query, asking for the  synonym/antonym of a target word through \ac{MLM} (e.g., ``happy is the synonym of [MASK]'').

Let $s_w$ and $a_w$ denote masked queries that ask the synonym and antonym of the word $w$, respectively. Also, let $A_s$ and $A_a$ refer to the list of correct answers for $s_w$ and $a_w$, respectively. Intuitively, $A_a$ becomes the wrong predictions of $s_w$, because $s_w$ and $a_w$ have the opposite meaning. Therefore, we can evaluate the violation of the \ac{LNP} by investigating whether a \ac{PLM} generates wrong predictions.

To extract commonly-used words for our experiment, we first extracted nouns, adjectives, and adverbs that appear more than five times in the SNLI dataset \cite{snli}. Among the extracted candidates, we filtered words that have synonyms or antonyms in ConceptNet. Finally, we generated  masked queries by employing templates used by \citet{camburu2020make}. As a result, we collected about 27K data points for  \ac{MWR}. The templates and examples of the data are  in Table~\ref{table.example_MWR} in Appendix~\ref{sec:example_appendix}.

\subsection{Synonym/Antonym Recognition}
 \ac{SAR} is a classification that distinguishes whether two given words are synonyms or antonyms. It aims to evaluate whether the contextualised representations of \acp{PLM} reflect the lexical meaning of words. Therefore, we use a parametric probing model \cite{adi2017fine, liu2019linguistic, belinkov2019analysis,sinha2021masked} for the experiment. Specifically, the experiment is performed on the final layer of each \acp{PLM}, i.e., we only train the classifier while keeping the encoder frozen. We use ConceptNet to build the dataset. ConceptNet has much more synonym triplets compared to antonyms. As a result, we randomly sample the synonym triplets to maintain a balance. To that end, we collect 33K, 1K, and 2K data points for the train, dev, and test datasets, respectively.

\subsection{Evaluation Metrics}
We use the top-$k$ hit rate (HR@$k$) to evaluate the performance on the \ac{MKR-NQ} and \ac{MWR} tasks. Assume that $P=\{(p_1, c_1), (p_2, c_2), \ldots, (p_n, c_n)\}$ denotes the set of predictions for a data point $x$, where $p_t$ and $c_t$ refer to the predicted word and confidence score of the $t$-th prediction, respectively. Then, the top-$k$ hit rate for a data point $x$ is defined as follows:
\begin{equation*}
    \mathit{HR}@k(x) = \frac{\sum_{i=1}^k \mathbbm{1}(p_i \in \mathcal{W}_x)}{k},
\end{equation*}
where $\mathcal{W}_x$ is the wrong prediction set of $x$. Intuitively, the metric measures the ratio of top-$k$ predicted words that belong to the wrong prediction set.

To reflect the prediction confidence score to the evaluation metric, we additionally define the weighted top-$k$ hit rate (WHR@$k$) that uses the confidence score as weights. It is worth to mention that lower metrics mean a better model performance in both cases as the metrics assess how likely the models make inaccurate answers that they must avoid. The weighted metric can be defined as follows:
\begin{equation*}
    \mathit{WHR}@k(x) = \frac{\sum_{i=1}^k c_i \times \mathbbm{1}(p_i \in \mathcal{W}_x)}{\sum_{i=1}^k c_i}.
\end{equation*}

For the \ac{SAR} task, we employ accuracy as an evaluation metric, because each data point has its own label, and the label distribution is not skewed.

% Table -------------------------------------------------------
\begin{table*}[t!]
	\begin{center}
		\renewcommand{\arraystretch}{1.1}
		\footnotesize{
			\centering{\setlength\tabcolsep{4pt}
		\begin{tabular}{c|ccccc|ccccc}
		\toprule
		\multirow{2}{*}{Model} & 
		\multicolumn{5}{c|}{MKR-NQ} & \multicolumn{5}{c}{MWR} \\ 
		& HR@1 & HR@3 & WHR@3 & HR@5 & WHR@5
		& HR@1 & HR@3 & WHR@3 & HR@5 & WHR@5
		\\ \hline
        
		BERT-\textit{base} 
		& 9.57 & 6.38 & 8.42 & 5.00 & 7.81
		& 35.03 & 18.83 & 28.71 & 13.26 & 26.03
		\\
		BERT-\textit{large} 
		& 13.33 & 7.70 & 11.17 & 6.03 & 10.51
		& 36.66 & 20.45 & 29.68 & 14.60 & 26.56
		\\ \hline

		RoBERTa-\textit{base} 
		& 11.52 & 6.85 & 9.63 & 5.30 & 8.91
		& 13.02 & 8.47 & 10.50 & 6.54 & 9.32
		\\
		RoBERTa-\textit{large} 
		& 15.72 & 9.25 & 13.31 & 6.86 & 12.24
		& 27.92 & 17.07 & 23.06 & 12.84 & 20.82
		\\ \hline
		
		ALBERT-\textit{base} 
		& 4.24 & 3.75 & 4.24 & 3.26 & 4.09
		& 26.37 & 14.95 & 22.10 & 10.75 & 20.32
		\\
		ALBERT-\textit{large} 
		& 9.22 & 6.38 & 7.96 & 4.94 & 7.30
		& 50.77 & 25.05 & 42.67 & 17.03 & 39.09
		\\
		\bottomrule
		\end{tabular}}}
	\vspace*{-1ex}
	\caption{Overall results for the \ac{MKR-NQ} and \ac{MWR} experiments. We multiply 100 to each value to improve readability. Note that the lower the values the better.}
	\vspace*{-1ex}
	\label{table.MKR_MWR_result}%
	\end{center}
	\vspace*{-1ex}
\end{table*}
% End Table ----------------------------------------------------

% Table -------------------------------------------------------
\begin{table}[t!]
	\begin{center}
		\renewcommand{\arraystretch}{1.1}
		\footnotesize{
			\centering{\setlength\tabcolsep{4pt}
		\begin{tabular}{c|cc|cc|cc}
		\toprule
		\multirow{2}{*}{} & 
		\multicolumn{2}{c|}{BERT} & \multicolumn{2}{c|}{RoBERTa} & \multicolumn{2}{c}{ALBERT} \\ 
		& \textit{base} & \textit{large}
		& \textit{base} & \textit{large}
		& \textit{base} & \textit{large}
		\\ \hline
		$\mathcal{R}_{syn}$
		& 37.79 & 36.58 & 17.13 & 46.01 & 11.37 & 76.10 \\
		$\mathcal{R}_{ant}$
		& 41.44 & 41.79 & 13.57 & 30.21 & 33.26 & 62.65 \\
		\bottomrule
		\end{tabular}}}
	\vspace{-1ex}
	\caption{Ratios of instances that \acp{PLM} regenerate the word in the input sentence. $\mathcal{R}_{syn}$ and $\mathcal{R}_{ant}$ are the ratios of synonym and antonym-asking questions, respectively.} 
	\vspace{-1ex}
	\label{table.ant_syn_ratio}%
	\end{center}
	\vspace{-1ex}
\end{table}
% End Table ----------------------------------------------------

\section{\texorpdfstring{\acp{PLM}}{} Lack Information of Negation and Lexical Semantics}\label{section.probing_exp}

We select the following \acp{PLM} for the experiments: \ac{BERT}-\textit{base/large} \cite{BERT}, RoBERTa-\textit{base/large} \cite{roberta}, and ALBERT-\textit{base/large} \cite{lan2019albert}. These \acp{PLM} are pre-trained with the \ac{MLM} training objective. We added the ELECTRA-\textit{small/base/large} models \cite{electra} for the \ac{SAR} task, but it is not used for the \ac{MKR-NQ} and \ac{MWR} experiments, as the discriminator of the ELECTRA models are trained with the \ac{RTP} training objective and have no \ac{MLM} classifier. No additional training is required for the \ac{MKR-NQ} and \ac{MWR} tasks. For the \ac{SAR} task, we fine-tune each \ac{PLM} for 10 epochs and apply the early stopping technique. We use the AdamW optimiser \cite{loshchilov2018fixing} for training with a learning rate of $5e^{-6}$ and a batch size of 32.

\subsection{Results for \texorpdfstring{\ac{MKR-NQ}}{}}
The results for the \ac{MKR-NQ} task are summarised in Table \ref{table.MKR_MWR_result}. In general, the results are consistent with previous works \cite{ettinger2020bert, kassner2020negated}. We observe three important characteristics from the experimental results.

First, large models produce a higher hit rate than their corresponding base-size models in all  three \acp{PLM}, recording an average of about 1.5 times higher values. This implies that large-size models are more likely to generate wrong predictions for negated queries, even though they perform better than small-size models in many benchmark tests. The results suggest that evaluating a model's performance solely based on the accuracy metric is unwise.

Second, the hit rate decreases as $k$ increases, which implies that the majority of \acp{PLM}' top predictions (e.g., $k$=1 or $k$=2) are incorrect. Finally, the weighted hit rate is much higher than the vanilla hit rate, suggesting that \acp{PLM} generate wrong predictions with high confidence.

% Table -------------------------------------------------------
\begin{table}[t!]
	\begin{center}
		\renewcommand{\arraystretch}{1.1}
		\footnotesize{
			\centering{\setlength\tabcolsep{4pt}
		\begin{tabular}{c|cc|cc}
		\toprule
		\multirow{2}{*}{Model} & 
		\multicolumn{2}{c|}{Encoder-fixed} & \multicolumn{2}{c}{Fine-tune} \\ 
		& $\mathcal{A}_{val}$ & $\mathcal{A}_{test}$
		& $\mathcal{A}_{val}$ & $\mathcal{A}_{test}$
		\\ \hline
		BERT-\textit{base} (108M) 
		& 53.1 & 55.0 & 84.0 & 85.6 \\
		BERT-\textit{large} (333M) 
		& 54.4 & 53.5 & 92.1 & 92.5 \\ \hline
    
		RoBERTa-\textit{base} (124M)
		& 71.1 & 70.1 & 87.2 & 87.8 \\
		RoBERTa-\textit{large} (355M)
		& 69.7 & 69.1 & 93.7 & 94.2 \\ \hline
		
		ALBERT-\textit{base} (11M)
		& 56.6 & 58.1 & 81.5 & 84.0 \\
		ALBERT-\textit{large} (17M)
		& 54.7 & 56.6 & 86.9 & 88.0 \\ \hline

		ELECTRA-\textit{small} (13M)
		& 64.1 & 63.9 & 80.2 & 80.9 \\
		ELECTRA-\textit{base} (109M)
		& 67.9 & 70.6 & 93.3 & 92.9 \\
		ELECTRA-\textit{large} (334M)
		& 69.4 & 72.7 & 95.9 & 95.4 \\
		\bottomrule
		\end{tabular}}}
	\vspace{-1ex}
	\caption{Results of the \ac{SAR} experiment. $\mathcal{A}_{val}$ and $\mathcal{A}_{test}$ are the accuracy of the \textit{validation} and \textit{test} dataset, respectively. We record the average of five repetitions.} 
	\vspace{-1ex}
	\label{table.SAR_result}%
	\end{center}
	\vspace{-1ex}
\end{table}
% End Table ----------------------------------------------------

\subsection{Results for \texorpdfstring{\ac{MWR}}{}}
The results of the \ac{MWR} task are summarised in Table~\ref{table.MKR_MWR_result}. The three characteristics found in the \ac{MKR-NQ} task are also observed in the \ac{MWR} task. Also, we found the following additional patterns.

\paragraph{\acp{PLM} lack knowledge of antonyms. } In general, the hit rates are extremely high compared to the \ac{MKR-NQ} task in all the \acp{PLM}. Analysing their predictions, we find that \acp{PLM} generate incorrect predictions primarily in antonym-asking queries. Specifically, the average HR@1 of the antonym-asking queries is 41.9\%, while that of the synonym-asking queries is only 1.4\%. A leading cause is that \acp{PLM} simply replicate the word presented in the input query. Table \ref{table.ant_syn_ratio} shows the ratio of instances where each \ac{PLM} reproduces the same word in a question. While the values are quite high for both synonym-asking and antonym-asking queries, the problem is more severe in the latter case, because the generated predictions are definitely incorrect. Based on our results, we conclude that \acp{PLM}' contextualised representations lack lexical semantic information. Our conclusion is in line with the findings of \citet{liu2019linguistic} showing that encoder-fixed \acp{PLM} are not suitable to deal with tasks that require fine-grained linguistic knowledge.

\paragraph{Issues are more severe with nouns. } We observe that the hit rates are higher when a word in a question is a noun. Specifically, the average HR@1 values of nouns, adjectives, and adverbs are 35.1\%, 27.4\%, and 11.8\%, respectively. Interestingly,  \acp{PLM} have a high error rate when dealing with nouns even though they are trained with a large written English corpus, where nouns form the greatest portion (at least 37\%) of all \ac{POS} tags~\cite{hudson199437, liang2013noun}.

\subsection{Results for \texorpdfstring{\ac{SAR}}{}}
As part of the comparison, we fine-tune each \ac{PLM} on the \ac{SAR} task, i.e., train the entire set of parameters. The results are summarised in Table \ref{table.SAR_result}. We observe a huge gap between the performance of fine-tuned models and that of encoder-fixed models. In contrast to the fine-tuned models that produce a high accuracy, encoder-fixed models fall short of expectations, even recording almost a random guess performance in \ac{BERT} models. Also, just as a common belief, large models' performance is greatly improved when fine-tuned. However, the difference between the large and small encoder-fixed models is insignificant, except for the ELECTRA models that exhibit only a marginal improvement. The two phenomenons suggest that \acp{PLM}' outstanding performance is predicated on updating  many parameters to learn syntactic associations presented in training data \cite{niven2019probing, mccoy2019right}, but their contextualised representations do not carry abundant lexical meaning information.

\section{Intermediate Training on Meaning Matching Task: \texorpdfstring{$\textrm{I}\textrm{M}^2$}{}}\label{sec:remedy_imm}
\subsection{Issue of \texorpdfstring{\acp{PLM}}{}}
Through the previous experiments, we observe that \acp{PLM} contain little information about negation and especially lexical semantics. We hypothesise a leading cause lies in the training objective of \acp{PLM}: the \ac{LM} objective, which is a backbone pre-training task of almost all \acp{PLM}.

In the \ac{LM} objective, words are generated based on given contexts. The \textit{distributional hypothesis} \cite{harris1954distributional}, which assumes that semantically related or similar words will appear in similar contexts \cite{mrksic2016counter}, is the underpinning assumption of the \ac{LM} objective \cite{sinha2021masked}. Under this assumption, a model learns the meaning of texts based on their correlation to others. This is a great benefit, because a model can learn the meaning of texts using only the text form, allowing unsupervised training. Based on this advantage, many unsupervised representations, such as Word2Vec \cite{word2vec}, Glove \cite{glove}, and current \acp{PLM}, have been developed.

However, the problem is that the \textit{distributional hypothesis} has limitations in reflecting a word's \textit{semantic} meanings, because words having different or even opposite semantic meanings can appear in similar or the same contexts. For instance, consider the two words ``boy'' and ``girl''. We can readily imagine sentences in which the two words appear in the same context, e.g., ``the little boy/girl cuddled the teddy bear closely''. As a result, a model can learn their common functional meanings, i.e., young human beings, and the vector representations would be very similar if they were trained based on the \textit{distributional hypothesis}. However, the representation hardly captures their semantic antonomy, e.g., gender. Similarly, negated sentences have almost identical contexts to their original forms. As a result, models cannot effectively learn the semantic meaning of words and negation expressions, provided they leverage only the text~forms.

% However, the problem is that the \textit{distributional hypothesis} does not consistently hold in natural language. Words having different meanings can appear in similar or even the same contexts. For instance, consider the two words ``boy'' and ``girl''. Despite their antonymy, we can readily imagine sentences in which the two words appear in the same context, e.g., ``The little boy/girl cuddled the teddy bear closely.''. As a result, the meaning of the two words would become quite similar if they were trained based on the \textit{distributional hypothesis}.\footnote{The cosine similarity between the Word2Vec vectors of ``boy'' and ``girl'' is 0.85 (and 0.74 for ``high'' and ``low''.)} Similarly, negated sentences have almost identical contexts with their original forms. As a result, models can not learn the true meaning of words and negation expressions, provided they leverage only the text forms.

% Table -------------------------------------------------------
\begin{table*}[t!]
	\begin{center}
		\renewcommand{\arraystretch}{1.1}
		\footnotesize{
			\centering{\setlength\tabcolsep{4pt}
		\begin{tabular}{c|ccccc|ccccc}
		\toprule
		\multirow{2}{*}{Model} & 
		\multicolumn{5}{c|}{MKR-NQ} & \multicolumn{5}{c}{MWR} \\ 
		& HR@1 & HR@3 & WHR@3 & HR@5 & WHR@5
		& HR@1 & HR@3 & WHR@3 & HR@5 & WHR@5
		\\ \hline
        
		BERT-\textit{large} 
		& 13.33 & 7.70 & 11.17 & 6.03 & 10.51
		& 36.66 & 20.45 & 29.68 & 14.60 & 26.56
		\\
		BERT-\textit{large} ($\textrm{I}\textrm{M}^2$)
		& \textbf{11.41} & \textbf{7.01} & \textbf{9.86} & \textbf{5.57} & \textbf{9.14}
		& \textbf{18.92} & \textbf{13.07} & \textbf{15.78} & \textbf{10.30} & \textbf{14.14}
		\\ \hline

		RoBERTa-\textit{large} 
		& 15.72 & 9.25 & 13.31 & 6.86 & 12.24
		& 27.92 & 17.07 & 23.06 & 12.84 & 20.82
		\\
		RoBERTa-\textit{large} ($\textrm{I}\textrm{M}^2$)
		& \textbf{6.56} & \textbf{4.97} & \textbf{6.05} & \textbf{3.99} & \textbf{5.67}
		& \textbf{22.08} & \textbf{12.68} & \textbf{18.94} & \textbf{9.20} & \textbf{17.63}
		\\
		
		\bottomrule
		\end{tabular}}}
	\vspace*{-1ex}
	\caption{Results of BERT-\textit{large} and RoBERTa-\textit{large} after applying the $\textrm{I}\textrm{M}^2$ approach. We multiply 100 to each value for a better readability. Note that the lower the values the better.}
	\vspace*{-1ex}
	\label{table.MM_MKR_MWR_result}%
	\end{center}
	\vspace*{-1ex}
\end{table*}
% End Table ----------------------------------------------------

% Table -------------------------------------------------------
\begin{table}[t!]
	\begin{center}
		\renewcommand{\arraystretch}{1.1}
		\footnotesize{
			\centering{\setlength\tabcolsep{2pt}
		\begin{tabular}{c|cc|cc}
		\toprule
		\multirow{2}{*}{Model} & 
		\multicolumn{2}{c|}{Encoder-fixed} & \multicolumn{2}{c}{Fine-tune} \\ 
		& $\Delta \mathcal{A}_{val}$ & $\Delta \mathcal{A}_{test}$
		& $\Delta \mathcal{A}_{val}$ & $\Delta \mathcal{A}_{test}$
		\\ \hline
		BERT-\textit{base} (108M) 
		& \textcolor{blue}{+5.5}* & \textcolor{blue}{+5.1}* & \textcolor{blue}{+3.9}* & \textcolor{blue}{+3.0}* \\
		BERT-\textit{large} (333M) 
		& \textcolor{blue}{+3.1}* & \textcolor{blue}{+6.3}* & \textcolor{blue}{+1.0} & \textcolor{blue}{+0.2} \\ \hline
    
		RoBERTa-\textit{base} (124M)
		& \textcolor{blue}{+4.5}* & \textcolor{blue}{+5.9}* & \textcolor{blue}{+1.3}* & \textcolor{blue}{+1.6}* \\
		RoBERTa-\textit{large} (355M)
		& \textcolor{blue}{+15.0}* & \textcolor{blue}{+17.1}* & \textcolor{blue}{+0.6} & \textcolor{blue}{+0.5} \\ \hline

		ALBERT-\textit{base} (11M)
		& \textcolor{red}{-2.6} & \textcolor{blue}{+2.5} & \textcolor{blue}{+4.7}* & \textcolor{blue}{+3.3}* \\
		ALBERT-\textit{large} (17M)
		& \textcolor{blue}{+1.3} & \textcolor{blue}{+1.4} & \textcolor{blue}{+1.2} & \textcolor{blue}{+1.6} \\ \hline

		ELECTRA-\textit{small} (13M)
		& \textcolor{red}{-4.1}* & \textcolor{red}{-2.7}* & \textcolor{blue}{+1.1} & \textcolor{blue}{+1.1} \\
		ELECTRA-\textit{base} (109M)
		& \textcolor{blue}{+3.8}* & \textcolor{blue}{+3.2}* & \textcolor{red}{-0.2} & \textcolor{blue}{+0.7} \\
		ELECTRA-\textit{large} (334M)
		& \textcolor{blue}{+14.0}* & \textcolor{blue}{+10.2}* & \textcolor{blue}{+0.4} & \textcolor{blue}{+0.5} \\
		\bottomrule

		\end{tabular}}}
	\vspace{-1ex}
	\caption{\acp{PLM}' accuracy change  in the \ac{SAR} task when we apply $\textrm{I}\textrm{M}^2$. We record the average across 5 runs. Our models show a statistically significant difference with $p$-value $<$ 0.05 (*) compared to the baseline results in Table~\ref{table.SAR_result}.}
	\vspace{-1ex}
	\label{table.MM_Result_SAR}%
	% s-level: 0.05
	\end{center}
	\vspace{-1ex}
\end{table}
% End Table ----------------------------------------------------

\subsection{Meaning-Matching Task}
In the light of meaning-text theory, there is a correspondence between linguistic expressions (\textit{text}) and semantic contents (\textit{meaning})~\cite{MTT1, MTT2}. Instead of solely relying on the \textit{distributional hypothesis}, we propose the new  \textit{meaning-matching} task, which can directly learn the correspondence. Specifically, meaning-matching is a classification  that takes a word and a sentence as input and determines whether the sentence defines the word correctly. Through this task, a model can learn both meaning-text correspondences and correlations between a word and other words in a definition, which is rarely found in general corpora.

For training \acp{PLM} on our new task, we apply the  \textit{intermediate-training} technique~\cite{phang2018sentence, wang2019tell, liu2019linguistic, pruksachatkun2020intermediate, vu2020exploring}, which first fine-tunes \acp{PLM} on an intermediate task, and then fine-tunes the model again on target tasks. It has been shown that training on intermediate tasks that require high-level linguistic knowledge and inference ability could improve performance~\cite{liu2019linguistic, pruksachatkun2020intermediate}. Furthermore, it is more efficient in time and resources than pre-training models on large corpora (e.g., BERTNOT model \cite{hosseini2021understanding}).  

\paragraph{Dataset. } We collect about 150K free-text definitions that depict the meaning of English words from \textbf{WordNet}~\cite{wordnet} and the \textbf{English Word, Meaning, and Usage Examples} dataset.\footnote{\href{https://data.world/idrismunir/english-word-meaning-and-usage-examples/}{https://data.world/idrismunir/english-word-meaning-and-usage-examples/}} In cases when a word appears in both datasets, we concatenate the word's definitions. Several examples of our data are presented in Table~\ref{table.example_meaning_matching} in Appendix~\ref{sec:example_appendix}. We use publicly available English datasets for convenience, but our approach is easily adaptable to other languages, since most of them have their own dictionaries.

\begin{figure}[ht]
	\centering
	\begin{subfigure}[b]{0.47\textwidth}
		\includegraphics[width=\linewidth]{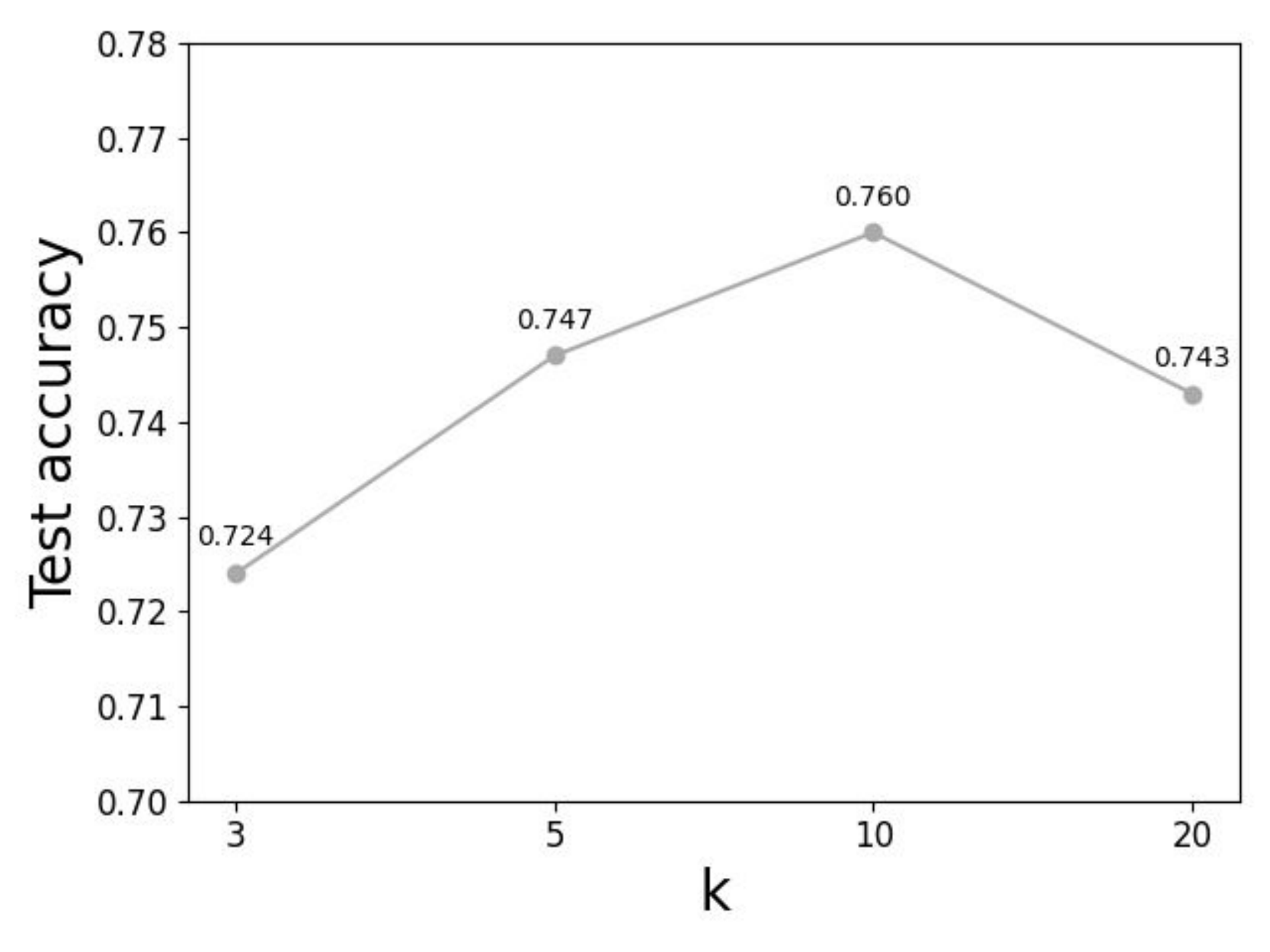}
	\end{subfigure}
	\vspace{-2ex}
	\caption{The performance of the RoBERTa-\textit{base} ($\textrm{I}\textrm{M}^2$) model with different $k$ values. We repeat each experiments for five times and record their average.}
	\vspace{-2ex}
    \label{fig:hps_sar}
\end{figure}
%% -----------------------------------------------------

\paragraph{Training details. } It is necessary to generate false word-definition pairs to train \acp{PLM} on the meaning-matching task. To achieve this, we use a negative sampling technique. We investigate the proper $k$ in the range of 3, 5, 10, and 20. For a hyperparameter search, the performance of the RoBERTa-\textit{base} model on the \ac{SAR} task is used as a criterion. Figure~\ref{fig:hps_sar} illustrates the \ac{SAR} performance of the RoBERTa-\textit{base} model with different $k$ values. Intuitively, a large $k$ value will lead the model to a better performance by investigating more word-meaning combinations. However, we observe that the model performs the best when $k$ is 10, and the performance decreases if $k$ is too large. We conjecture that a leading cause is that the dataset contains many words with similar meanings, mostly derived from the same \textit{stem}. As a result, large $k$ values can increase the possibility of recognising the meaning of such similar words as different.

% We set the number of negative samples $k$ to 10. More details regarding determining the value are described in Appendix \ref{appendix.n_neg}. 

To avoid the class-imbalance issue in a batch, we duplicate the correct word-definition pairs $k$ times when we construct the training data. For training, the AdamW optimiser is used with a learning rate of $5e^{-6}$. We use 5\% of data points for validation and train the models for 15 epochs with a batch size of 32. The early stopping technique is used to prevent overfitting.

\section{Experiments and Results}
We conduct the same probing tasks after the intermediate training on the meaning-matching task.\footnote{Our models trained with the meaning-match task can be downloaded from the following repositories: \href{https://huggingface.co/korca/meaning-match-electra-large}{ELECTRA-large}, \href{https://huggingface.co/korca/meaning-match-bert-large}{BERT-large}, \href{https://huggingface.co/korca/meaning-match-roberta-large}{RoBERTa-large}.}

\begin{table}
	\begin{center}
		\renewcommand{\arraystretch}{1.1}
		\footnotesize{
			\centering{\setlength\tabcolsep{2pt}
				\begin{tabular}{p{35mm}|p{35mm}}
					\toprule
					%1
                    \multicolumn{2}{c}{\makecell[l]{
                    \textsc{Query}: demand is an antonym of [MASK]}}
                    \\ \midrule
                    \makecell[c]{\textsc{RoBERTa-large}} & \makecell[c]{\textsc{RoBERTa-large ($\textrm{I}\textrm{M}^2$)}} \\
                    \makecell[c]{demand} & \makecell[c]{supply}\\
                    \midrule
					%2
                    \multicolumn{2}{c}{\makecell[l]{
                    \textsc{Query}: tomorrow is the opposite of [MASK]}}
                    \\ \midrule
                    \makecell[c]{\textsc{BERT-large}} & \makecell[c]{\textsc{BERT-large ($\textrm{I}\textrm{M}^2$)}} \\
                    \makecell[c]{tomorrow} & \makecell[c]{today}\\
                    \midrule
					%3
                    \multicolumn{2}{c}{\makecell[l]{
                    \textsc{Query}: question is an antonym of [MASK]}}
                    \\ \midrule
                    \makecell[c]{\textsc{BERT-large}} & \makecell[c]{\textsc{BERT-large ($\textrm{I}\textrm{M}^2$)}} \\
                    \makecell[c]{question} & \makecell[c]{answer}\\
                    \bottomrule	
		\end{tabular}}}
	\vspace{-1ex}
	\caption{Examples of top-1 predictions on \ac{MWR} queries. Unlike the original \acp{PLM}, our models do not reproduce a word in a query and make quite accurate predictions.}
	\vspace{-1ex}
	\label{table.example_MM_MWR}%
	\end{center}
    \vspace{-2ex}
\end{table}

\subsection{\texorpdfstring{\ac{SAR}}{} Results} \label{sec:mm_sar}
We first focus on the \ac{SAR} task. After the intermediate training, all models are fine-tuned on the \ac{SAR} task with the same hyperparameters described in Section~\ref{section.probing_exp}. The results are summarised in Table~\ref{table.MM_Result_SAR}.

\paragraph{Improved lexical semantic information. } We generally observe marginal or no significant improvements when fine-tuning the whole parameters, especially for large-size \acp{PLM}. However, with fixed encoder, the performance is significantly improved for \acp{PLM} with more than 100M parameters, and the improvements are more significant for large \acp{PLM}. Our results show that the proposed approach assists \acp{PLM} to learn enhanced representations with more abundant lexical semantic information.

%%%%%%%%%%%%%%%%%%%%%%%%%%%%%%%%%%%%%%%%%%%%%%%%%%%%%%%%%%%%
\begin{figure}[t!]
	\centering
	\begin{subfigure}[b]{0.5\textwidth}
		\includegraphics[width=\linewidth]{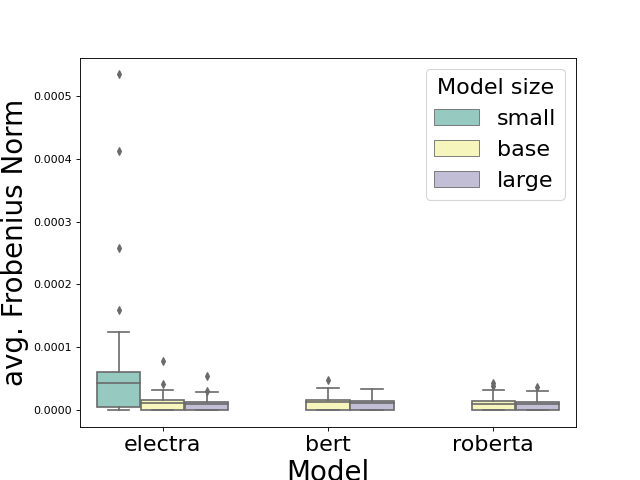}
	\end{subfigure}
	\vspace{-2ex}
	\caption{Frobenius norm box plots of \acp{PLM}' layer after intermediate training on the meaning-matching task.}
	\vspace{-2ex}
	\label{fig:model_diff}
\end{figure}
%% -----------------------------------------------------

\begin{table*}[t!]
	\begin{center}
		\renewcommand{\arraystretch}{1.1}
		\footnotesize{
			\centering{\setlength\tabcolsep{3pt}
		\begin{tabular}{c|cccccccc}
		\toprule
		Model & 
		COLA & MNLI-m & MNLI-mm & QNLI & RTE & QQP & MRPC & SST2 \\ \hline
		BERT-\textit{large}
		& 59.6$\pm$1.1 & 85.5$\pm$0.4 & 85.3$\pm0.5$ & \textbf{91.7}$\pm$0.1 & 65.5$\pm$2.6  & 89.9$\pm$0.2 & 80.9$\pm$2.0 & 92.3$\pm$0.3 \\
		BERT-\textit{large} ($\textrm{I}\textrm{M}^2$)
		& \textbf{61.5}$\pm$1.0& \textbf{85.7}$\pm$0.1 & \textbf{85.5}$\pm$0.1 & 91.6$\pm$0.2 & \textbf{66.8}$\pm$1.0 & \textbf{90.0}$\pm$0.1 & \textbf{82.8}$\pm$1.1 & \textbf{92.4}$\pm$0.3 \\ \hline
		
		RoBERTa-\textit{large}
		& 62.9$\pm$1.9& 90.2$\pm$0.1 & \textbf{90.0}$\pm$0.2 & \textbf{94.5}$\pm$0.1 & 81.7$\pm$1.8
		& 90.9$\pm$0.4 & 87.2$\pm$1.1 & \textbf{95.7}$\pm$0.1 \\
		RoBERTa-\textit{large} ($\textrm{I}\textrm{M}^2$)
		& \textbf{64.8}$\pm$2.1& \textbf{90.3}$\pm$0.1 & 89.9$\pm$0.1 & 94.4$\pm$0.1 & \textbf{83.1}$\pm$1.4 & \textbf{91.0}$\pm$0.0 & \textbf{88.2}$\pm$1.5 & 95.4$\pm$0.3 \\ \hline
		
		ELECTRA-\textit{large}
		& 68.4$\pm$2.3& \textbf{90.9}$\pm$0.1 & 90.7$\pm$0.2 & \textbf{94.5}$\pm$0.3 & 86.9$\pm$2.2
		& 91.6$\pm$0.5 & 88.9$\pm$1.5 & \textbf{96.7}$\pm$0.1 \\
		ELECTRA-\textit{large} ($\textrm{I}\textrm{M}^2$)
		& \textbf{69.1}$\pm$0.7& 90.8$\pm$0.1 & \textbf{90.7}$\pm$0.1 & 94.3$\pm$0.2 & \textbf{87.0}$\pm$1.3 & \textbf{91.7}$\pm$0.3 & \textbf{89.5}$\pm$0.5 & 96.4$\pm$0.4 \\

		\bottomrule
		\end{tabular}}}
	\vspace*{-1ex}
	\caption{GLUE benchmark validation performance of \acp{PLM} before and after intermediate training on the meaning-matching task. Matthew's correlation for the COLA and accuracy for the other tasks are used as an evaluation metric. We report the mean and standard deviation across 5 runs. The best values for each \ac{PLM} are in bold.} 
	\vspace*{-1ex}
	\label{table.glue_experiment}%
	\end{center}
	\vspace*{-1ex}
\end{table*}

\paragraph{Catastrophic forgetting. } We find that small \acp{PLM}, such as ELECTRA-\textit{small} and ALBERT models, show no significant increase in performance or are negatively impacted. Because all \acp{PLM} achieve a comparable performance on the meaning-matching task, we hypothesise that a leading cause is \textit{catastrophic forgetting} \cite{pruksachatkun2020intermediate, singh2020bertnesia}, where the model forgets previous knowledge learned through pre-training to accept new information from the intermediate task. To verify this, we measure the change of parameter values after $\textrm{I}\textrm{M}^2$. Concretely, let $M_i$ and $M^{mm}_i$ denote the parameter of $i$-th layer before and after $\textrm{I}\textrm{M}^2$. We calculate the average Frobenius norm for each layer:
\begin{equation*}
    \mathcal{F}_i=\frac{1}{|M_i|} |M_i - M^{mm}_i|_F.
\end{equation*}

Figure~\ref{fig:model_diff} shows the boxplots of $\mathcal{F}_i$ for each \acp{PLM}. We observe that the parameters of the ELECTRA-\textit{small} model, which is negatively impacted, are changed considerably compared to other \acp{PLM} having parameters more than 100M. The results suggest that the size of \acp{PLM} is an important property to prevent the catastrophic forgetting issue.

\subsection{\texorpdfstring{\ac{MKR-NQ}}{} and \texorpdfstring{\ac{MWR}}{} Results}\label{sec:mm_mkr_mwr}
Next, we perform the \ac{MKR-NQ} and \ac{MWR} tasks after applying the $\textrm{I}\textrm{M}^2$ method. Since our models are not trained with the \ac{MLM} objective, we replace the encoder of original \acp{PLM} with that of the models after fine-tuning on the meaning-matching task and reuse the \ac{MLM} classifier. For the experiments, we use BERT-\textit{large} and RoBERTa-\textit{large}, because they are pre-trained based on the \ac{MLM} objective, and parameters are hardly changed after applying the $\textrm{I}\textrm{M}^2$ method. The results are summarised in Table~\ref{table.MM_MKR_MWR_result}.

We observe substantial decreases in the hit rates of incorrect predictions in both \acp{PLM}. For the \ac{MWR} task, we find that the issue of regenerating a word in a given query is greatly relieved after applying the $\textrm{I}\textrm{M}^2$ method. Specifically, the percentage of such instances drops from 40.3\% to 19.6\% and from 33.8\% to 25.2\% for BERT-\textit{large} and RoBERTa-\textit{large}, respectively. Several examples of the predicted results are presented in Table~\ref{table.example_MM_MWR}. The results lend support to our claim that the $\textrm{I}\textrm{M}^2$ approach is of benefit to learning lexical semantic information and the meaning of negated expressions.

\subsection{Fine-Tuning on the GLUE Benchmark}\label{sec:glue_exp}
A critical drawback of intermediate training is that the target task performance could be negatively impacted if the intermediate task is not related to the target task \cite{liu2019linguistic, pruksachatkun2020intermediate}. To confirm whether the issue occurs, we compare the performance of BERT, RoBERTa, and ELECTRA-\textit{large}  on 7 GLUE benchmark datasets \cite{GLUE} with their $\textrm{I}\textrm{M}^2$ counterparts. We train the models for 10 epochs for each dataset and apply the early stopping technique where the patience number is set to 3. It is observed that the training is generally finished within 8 epochs for all the models. The batch size per GPU and learning rates used for each dataset are described in Table~\ref{table.glue_param}. Datasets with large training set (e.g., MNLI, QNLI, and QQP) were not sensitive to the hyperparameters.

\begin{table}[t]
	\begin{center}
		\renewcommand{\arraystretch}{1.1}
		\footnotesize{
			\centering{\setlength\tabcolsep{2pt}
		\begin{tabular}{c|c|c|c|c|c|c|c}
		\toprule
		\multirow{2}{*}{} & 
		COLA & MNLI & QNLI & RTE & QQP & MRPC & SST2 \\ \hline
		b-size & 16 & 64 & 64 & 8 & 64 & 8 & 64 \\
		lr & $2e^{-5}$ & $1e^{-5}$ & $1e^{-5}$ & $2e^{-5}$ & $1e^{-5}$ & $1e^{-5}$ & $1e^{-5}$ \\
		\bottomrule
		\end{tabular}}}
	\vspace{-1ex}
	\caption{Batch size and learning rates used for the GLUE benchmark experiments.}
	\vspace{-1ex}\label{table.glue_param}%
	\end{center}
	\vspace{-1ex}
\end{table}

The results are presented in Table \ref{table.glue_experiment}. We find no significant difference in performance for tasks with large datasets, such as MNLI, QNLI, QQP, and SST2. On the contrary, tasks with small datasets, like MRPC and RTE, are slightly improved. The result is consistent with \citet{pruksachatkun2020intermediate} and \citet{vu2020exploring}, which showed that smaller tasks benefit much more from the intermediate training. Furthermore, unlike the previous studies that observed a negative transfer with the COLA dataset \cite{phang2018sentence, pruksachatkun2020intermediate}, the performance is improved in our approach. The result suggests that meaning-matching is a safe intermediate task that ensures a positive transfer with target downstream tasks.

% Table -------------------------------------------------------
\begin{table}[t!]
	\begin{center}
		\renewcommand{\arraystretch}{1.1}
		\footnotesize{
			\centering{\setlength\tabcolsep{2pt}
		\begin{tabular}{c|cc|cc}
		\toprule
		\multirow{2}{*}{Model} & 
		\multicolumn{2}{c|}{SNLI} & \multicolumn{2}{c}{MNLI} \\ 
		& dev & w/neg
		& dev & w/neg
		\\ \hline
        
		BERTNOT
		& 89.0$\pm$0.1 & 46.0$\pm$0.4 & \textbf{84.3}$\pm$2.3 & 60.9$\pm$0.3
		\\
		BERT-$\textrm{I}\textrm{M}^2$
		& \textbf{90.3}$\pm$0.2 & \textbf{48.00}$\pm$0.5 & 83.1$\pm$0.3 & \textbf{61.8}$\pm$0.6
		\\
		\bottomrule
		\end{tabular}}}
	\vspace{-1ex}
	\caption{Accuracies on the original development dataset (dev) and the NegNLI (w/neg) dataset for SNLI and MNLI tasks. The results of our approach are averaged across 5 runs. The best values are in bold.}
	\vspace{-1ex}\label{table.NegNLI_exp}%
	\end{center}
	\vspace{-1ex}
\end{table}
% End Table ----------------------------------------------------

\subsection{Experiments on the NegNLI Dataset}\label{sec:negnli_exp}
Finally, we conduct experiments on the NegNLI benchmark dataset \cite{hossain2020analysis},  where negation plays an important role for \ac{NLI} tasks. As a baseline, we compare the reported performance of BERTNOT  \cite{hosseini2021understanding}, which is a recently proposed remedy to improve \acp{PLM}' ability to understand negation. Since \citet{hosseini2021understanding} used BERT-\textit{base} as a backbone model, we also apply the $\textrm{I}\textrm{M}^2$ method to BERT-\textit{base}. The results are summarised in Table~\ref{table.NegNLI_exp}. 

For both SNLI and MNLI, we observe that our approach outperforms BERTNOT in the NegNLI datasets, while yielding a comparable performance in the original development datasets. It is interesting that our approach improves the understanding of negation in both \ac{MKR-NQ} and NegNLI tasks. We conjecture that a leading cause is that the definitions of the meaning-matching dataset contain many negation expressions, which enables a model to learn their proposed meaning (see Table~\ref{table.example_meaning_matching}). The results suggest that our proposed approach is more efficient than BERTNOT, because the $\textrm{I}\textrm{M}^2$ method leverages less time and resources for training.

\section{Related Work}
\label{sec:related_work}

\acp{PLM} are at the core of many success stories in  \ac{NLP}.
However, it remains unclear to what extent \acp{PLM} understand the syntactic and semantic properties of the human language.\
A series of probing tasks have been conducted on \acp{PLM} and have found them 
lacking or falling short on some language properties.
Among the many findings of these probing tasks, \acp{PLM} have been found to be
insensitive to the order of sentences when generating representations~\cite{%
pham2020out, %
gupta2021bert, %
sinha2021masked%
},
struggle to comprehend number-related representations
~\cite{wallace2019nlp, lin2020birds, nogueira2021investigating}, 
and  display a lack of semantic content understanding
~\cite{ravichander2020systematicity, elazar2021measuring}.

In addition to the above faulty behaviours,~\citet{ettinger2020bert} and
\citet{kassner2020negated} show that \acp{PLM} fail to comprehend \textit{negation},
which is an important property of language in many \ac{NLU} tasks.
\citet{ettinger2020bert} check the ability of \acp{PLM} to understand the meaning 
of negation in given contexts. 
In their work, they check whether models are sensitive in their completions of sentences
that either include \textit{negation} or not.
Under normal circumstances, the completions are expected to vary in truth depending on the 
presence or absence of negation in given sentences.
Their results show that \acp{PLM} are insensitive to the impacts of negations when
completing sentences.
\citet{kassner2020negated} construct the negated LAMA dataset by inserting
negation elements (e.g., ``not'') in the LAMA cloze questions \cite{petroni2019language}.
They use negated and original question pairs to query \acp{PLM} and establish that 
models are equally prone to make the same predictions for both the original and negated 
questions.
In a well-informed setting, it is expected that \acp{PLM} should make different
predictions for the original and negated questions.
This shows that \acp{PLM} struggle to comprehend negation.

In light of the highlighted faulty behaviours of \acp{PLM}, especially their struggle to
comprehend negation,~\citet{hosseini2021understanding} propose a remedy to alleviate the
problem.
In their remedy, they augment the language modelling objective with an unlikelihood
objective~\cite{welleck2019neural} based on negated sentences from the training corpus.
They use a syntactic augmentation method to generate negated sentences. 
In this method, the dependency parse of the sentences, \ac{POS} tags, and morphological information 
of each word  are taken as input, and the negation of sentences is done using sets of 
dependency tree regular expression patterns, such as Semgrex~\cite{Semgrex}. 
During training, they replace objects in negated sentences with [MASK] tokens and 
use unlikelihood training to make the masked-out tokens unlikely under the \ac{PLM} distribution. 
To ensure that negated sentences are factually false, they use the corresponding positive sentences as 
context for the unlikelihood prediction task.

Previous studies (e.g., \citet{kassner2020negated}) have mostly limited the scope of the logical 
negation property only to the negation expressions (e.g., ``no'' and ``not''). 
However, the core spirit of this property is the \textit{opposite meaning}, which is not only 
limited to the \textit{negation}. 
\citet{welleck2019neural} consider negating sentences using dependency tree regular expression
patterns.
This widens the scope of negation, as it is not only limited to the negation expressions ``no'' and ``not''.
However, their approach relies on other components, such as Semgrex, and  dependency and \ac{POS} parsers,  
which could impact the quality of the data, hence impact the models' performance.
In this work, we consider other perturbation methods to generate the opposite-meaning sentences 
to investigate whether \acp{PLM} satisfy the logical negation property,  and we propose a remedy,  
called \textbf{i}ntermediate-training on \textbf{m}eaning-\textbf{m}atching ($\textrm{I}\textrm{M}^2$), which hardly 
employs additional linguistic components.

\section{Summary and Outlook}
In this work, we investigated \acp{PLM}' \ac{LNP}. Compared to  previous works that only examine negation expressions, we expanded the boundary of \ac{LNP} to lexical semantics. We confirmed that \acp{PLM} are likely to violate \ac{LNP} through extensive experiments.

We hypothesise that the distributional hypothesis is an insufficient basis for understanding the semantic meaning of texts. To alleviate the issue, we proposed a novel intermediate task: meaning-match\-ing. Via  experiments, we verified that  meaning-matching is a stable intermediate task that substantially improves \acp{PLM}' understanding of negation and lexical semantic information while guaranteeing a positive transfer with multiple downstream tasks. Also, our approach produces a better performance on the negated \ac{NLI} datasets compared to the unlikelihood training-based method, which leverages much more time and resources. Our work suggests that it is time to move beyond the distributional hypothesis to develop logically consistent and stable language models.

% Entries for the entire Anthology, followed by custom entries
% ############ Acronyms ####################
\begin{acronym}
    \acro{LNP}{logical negation property}
    \acro{BERT}{bidirectional encoder representations from transformers}
    \acro{BiLSTM}{bidirectional LSTM}
    \acro{DL}{deep learning}
    \acro{KG}{knowledge graph}
    \acro{MLM}{masked language modelling}
    \acro{NLI}{natural language inference}
    \acro{NLP}{natural language processing}
    \acro{NLU}{natural language understanding}
    \acro{NN}{neural network}
    \acro{PLM}{pre-trained language model}
    \acro{SOTA}{state-of-the-art}
    \acro{POS}{parts of speech}
    \acro{MKR-NQ}{masked knowledge retrieval on negated queries}
    \acro{MWR}{masked word retrieval}
    \acro{SAR}{synonym/antonym recognition}
    \acro{RTP}{replaced token prediction}
    \acro{LM}{language modelling}
    \acro{QA}{question answering}
    \acro{STS}{semantic textual similarity}
\end{acronym}

\section*{Acknowledgements}
This work was partially supported by the Alan Turing Institute under the EPSRC grant EP/N510129/1, by the AXA Research Fund, by the EPSRC grant EP/R013667/1, 
and by the ESRC
grant ``Unlocking the Potential of AI for English Law''. We also acknowledge the use of Oxford’s ARC facility, of the EPSRC-funded Tier 2 facilities JADE (EP/P020275/1) and JADE~\Romannum{2}~(EP/T022205/1), and of GPU computing support by Scan Computers International Ltd.
Frank Mtumbuka was supported by the Rhodes Trust under a Rhodes Scholarship.

%\clearpage
\bibliography{reference}

\clearpage
\onecolumn
\appendix

\section{Appendix: Examples}\label{sec:example_appendix}

\begin{table*}[ht]
	\begin{center}
		\renewcommand{\arraystretch}{1.1}
		\footnotesize{
			\centering{\setlength\tabcolsep{2pt}
		\begin{tabular}{ccc}
		\toprule
        \textbf{Relation} & \textbf{Negated Query} & \textbf{Wrong Predictions} \\ \midrule
        \textit{IsA} & Truth isn't a [MASK]. & [``fact'', ``statement'', ``concept'', ``actuality''] \\ 
        \textit{CapableOf} & A doctor cannot [MASK] you. & [``care''] \\ 
        \textit{PartOf} & England isn't part of the [MASK]. & [``Europe''] \\ 
        \textit{HasA} & Apples don't have [MASK] inside them. & [``stems'', ``seeds''] \\ 
        \textit{UsedFor} & A map isn't for [MASK]. & [``navigate'', ``locating'', ``navigating'', ``orienteering'', ``information''] \\ 
        \textit{MadeOf} & Air doesn't have [MASK]. & [``molecules''] \\ 
        \textit{NotDesires} & Soldier does want to be [MASK]. & [``die''] \\
        \bottomrule
        \end{tabular}}}
    \vspace*{-1ex}
	\caption{ConceptNet relations for constructing the \ac{MKR-NQ} dataset and their corresponding sample data points.}
% 	\vspace*{-2ex}
	\label{table.example_MKR_NQ}%
	\end{center}
    % \vspace*{-2ex}
\end{table*}

\begin{table*}[ht]
	\begin{center}
		\renewcommand{\arraystretch}{1.1}
		\footnotesize{
			\centering{\setlength\tabcolsep{1pt}
		\begin{tabular}{c@{\ \ \ }c@{\ \ \ }c}
		\toprule
        \textbf{Template} & \textbf{Query} & \textbf{Wrong Predictions} \\ \midrule
        \textit{X is a synonym of Y} & boy is a synonym of [MASK]. & [``sister'', ``girl''] \\
        \textit{X is an antonym of Y} & boy is an antonym of [MASK]. & [``boys'', ``brat'', ``man'', ``boy'', ``lad'', \ldots]\\

        \textit{X is another form of Y} & learning is another form of [MASK]. & [``forgetting'', ``teaching''] \\
        \textit{X is the opposite of Y} & learning is the opposite of [MASK]. & [``knowledge'', ``erudition'', ``eruditeness'', ``learning'']\\

        \textit{X is a rephrasing of Y} & speaker is a rephrasing of [MASK]. & [``microphone'', ``listener'', ``addressee''] \\
        \textit{X is different from Y} & speaker is different from [MASK]. & [``loudspeaker'', ``transducer'', ``talker'', ``speaker'', \ldots]\\
        
        \bottomrule
        \end{tabular}}}
    \vspace*{-1ex}
	\caption{Templates used to construct the \ac{MWR} dataset and their sample data points.} 
    % \vspace*{-2ex}
    \label{table.example_MWR}%
	\end{center}
% 	\vspace*{-2ex}
\end{table*}

\begin{table*}[ht]
	\begin{center}
		\renewcommand{\arraystretch}{1.1}
		\footnotesize{
			\centering{\setlength\tabcolsep{2.4pt}
		\begin{tabular}{c@{\ \ \ }l}
		\toprule
        \textbf{Word} & \makecell[c]{\textbf{Definition}} \\ \midrule
        \textit{abnormal} & not normal; not typical or usual or regular or conforming to a norm; out of ordinary; unusual \\
        \textit{afebrile} & having no fever \\
        \textit{barefaced} & with no effort to conceal \\
        \textit{career} & \makecell[l]{the particular occupation for which you are trained; a job or occupation that a person does for an \\ extended period} \\
        \textit{cargo} & goods carried by a large vehicle \\
        \textit{revise} & the act of rewriting something; to review, alter and amend, especially of written material \\
        \textit{salary} & \makecell[l]{something that remunerates; a determined yearly amount of money paid to an employee by an employer \\ during a job} \\
        \bottomrule
        \end{tabular}}}
    \vspace*{-1ex}
	\caption{Examples of word-definition pairs that we used for the meaning-matching task.} 
% 	\vspace*{-2ex}
	\label{table.example_meaning_matching}%
	\end{center}
% 	\vspace*{-2ex}
\end{table*}

\end{document}